\documentclass[10pt,twocolumn,letterpaper]{article}

\usepackage{cvpr}              

\usepackage{booktabs}

\usepackage{times}
\usepackage{epsfig}
\usepackage{graphicx}
\usepackage{amsmath}
\usepackage{amssymb}

\usepackage{multirow}
\usepackage{subfloat}
\usepackage{color}
\usepackage[pagebackref=true,breaklinks=true,colorlinks,bookmarks=false]{hyperref}

\def\Dset{\mathcal{D} }
\def\Tset{\mathcal{T} }
\def\Pset{\mathcal{P} }
\def\Qset{\mathcal{Q} }
\def\Lset{\mathcal{L} }

\def\x{\mathbf{x}}
\def\I{\mathbf{I}}
\def\y{\mathbf{y}}

\def\C{\mathbf{C}}

\usepackage[accsupp]{axessibility}

\usepackage[capitalize]{cleveref}

\def\cvprPaperID{*****} 
\def\confName{CVPR}
\def\confYear{2022}

\makeatletter
\def\thanks#1{\protected@xdef\@thanks{\@thanks
        \protect\footnotetext{#1}}}
\makeatother

\begin{document}

\title{No-Reference Point Cloud Quality Assessment via Domain Adaptation}

\author{Qi Yang\textsuperscript{*}, Yipeng Liu\textsuperscript{*}, Siheng Chen, Yiling Xu, Jun Sun\\
Cooperative Medianet Innovation Center, Shanghai Jiao Tong University\\
{\tt\small \{yang\_littleqi, liuyipeng, sihengc, yl.xu, junsun\}@sjtu.edu.cn}
\thanks{$*$ equal contribution}
}
\maketitle

\begin{abstract}
We present a novel no-reference quality assessment metric, the image transferred point cloud quality assessment (IT-PCQA), for 3D point clouds.  For quality assessment, deep neural network (DNN) has shown compelling performance on no-reference metric design. However, the most challenging issue for no-reference PCQA is that we lack large-scale subjective databases to drive robust networks. Our motivation is that the human visual system (HVS) is the decision-maker regardless of the type of media for quality assessment. Leveraging the rich subjective scores of the natural images, we can quest the evaluation criteria of human perception via DNN and transfer the capability of prediction to 3D point clouds. In particular, we treat natural images as the source domain and point clouds as the target domain, and infer point cloud quality via unsupervised adversarial domain adaptation. To extract effective latent features and minimize the domain discrepancy, we propose a hierarchical feature encoder and a conditional-discriminative network. Considering that the ultimate purpose is regressing objective score, we introduce a novel conditional cross entropy loss in the conditional-discriminative network to penalize the negative samples which hinder the convergence of the quality regression network. Experimental results show that the proposed method can achieve higher performance than traditional no-reference metrics, even comparable results with full-reference metrics. The proposed method also suggests the feasibility of assessing the quality of specific media content without the expensive and cumbersome subjective evaluations. Code is available at https://github.com/Qi-Yangsjtu/IT-PCQA.
\end{abstract}
\section{Introduction}\label{section:introduction}
Point clouds have achieved compelling performance in augmented reality~\cite{lim2020Augmented}, automatic driving~\cite{ChenLFGW:20} and industrial robots~\cite{Rusu2011PCL}. Accurate point cloud quality assessment (PCQA) is a critical safeguard for providing high-quality point clouds for human vision tasks, such as virtual/augmented reality and cultural heritage. However, point clouds are subject to more complex distortion compared with image/video due to their data format, which means PCQA is a more challenging task. Specifically, point clouds consist of scattered points with spatial coordinates and attributes (e.g., color, normal, and opacity).
To ameliorate the urgent demand of PCQA, MPEG first adopts point-to-point/plane (p2point, p2plane) ~\cite{Mekuria2016Evaluation,tian2017geometric} to quantify point clouds geometrical distortion, and suggests to use the peak signal noise ratio (PSNR) of luminance and chrominance to deduce color distortion \cite{torlig2018novel}. After that, more reliable and robustness metrics are proposed, such as PCQM \cite{meynet2020pcqm}, GraphSIM \cite{yang2020inferring}, MPED \cite{yang2020MPED} and MS-GraphSIM \cite{Zhang2021MSGRAPHSIM}. Despite the emergence of these metrics alleviating the urgent demand of PCQA, these metrics are all full-reference metrics, which means the final evaluation results need the assistance of reference samples.

\begin{figure}[pt]
\setlength{\abovecaptionskip}{0.cm}
\setlength{\belowcaptionskip}{-0.cm}
	\centering
		\includegraphics[width=1\linewidth]{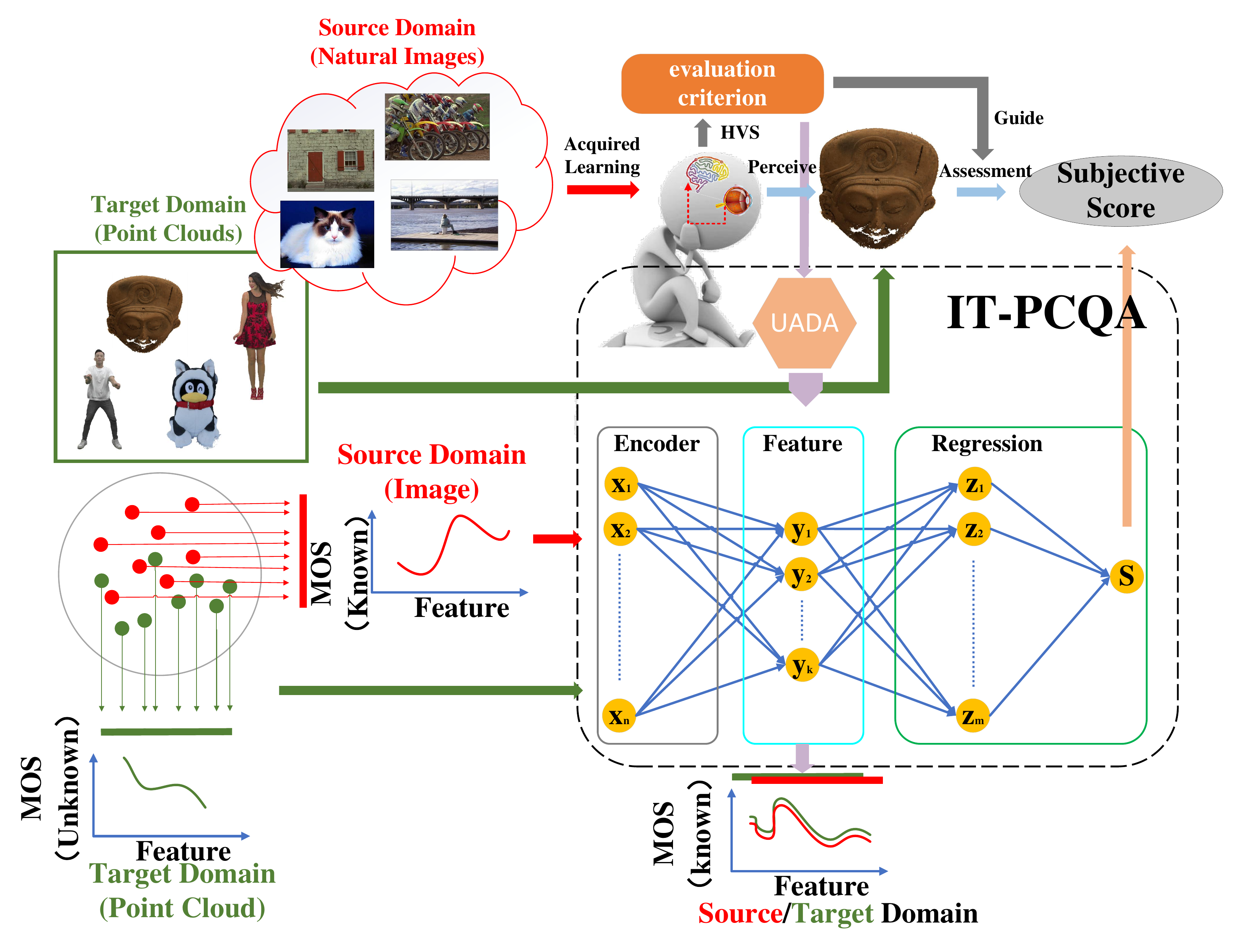}
	\caption{Motivation of domain adaption in PCQA. The source domain (natural images) and target domain (point clouds) have different domain distribution. Refer to the evaluation criterion specified by HVS, we use unsupervised adversarial domain adaptation (UADA)  to map the domain distribution and regress quality score. }
	\label{fig:uada}
\end{figure}

 However, in many practical applications, reference samples are not available, such as point clouds rendered after the transmission system \cite{LQrate}. And some samples naturally do not have a high-quality reference, such as the point clouds captured in the wild. Therefore, a more urgent task is studying PCQA in a no-reference case. Refer to the development of IQA, a no-reference metric usually based on natural scene statistic \cite{scstmm} or convolutional neural network (CNN)~\cite{dnntip}. Both methods need to analyze point cloud characteristics based on a large number of samples. But current PCQA databases are usually small, which are insufficient to derive a robust no-reference model. Therefore, the challenge of no-reference PCQA lies in lacking a complete subjective database with accurate labels.



 Therefore, in the absence of a large-scale subjective experiment, we propose to solve no-reference PCQA whereby unsupervised adversarial domain adaptation (UADA). Domain adaptation (DA) can reduce the need for costly labeled data in the target domain via using the rich samples and labels in the source domain \cite{DomainSurvey}. Considering the natural image quality assessment (IQA) research is relatively mature, we treat natural images as the source domain and point clouds as the target domain, and infer point cloud quality via inheriting the prior knowledge in the image field.

 Specifically, there are two reasons behind that. First,  there are numerous IQA databases \cite{LIVE}\cite{tid2013} with accurate subjective ratings, and the scale of IQA databases is much larger than that of PCQA.
Second, based on previous researches, there are strong connections between 2D and 3D perception, such as reconstruction (e.g., from image to 3D object~\cite{2D3Dre}) and tracking ~\cite{2D3Dtracking}. For quality assessment, our human visual system (HVS) is the universal evaluator, which means the perception characteristics shown in IQA share possible homogeneity with that in PCQA. Refer to Fig. \ref{fig:uada}, the characteristics of the natural image (i.e., the red curve) and point cloud (i.e., the green curve) have their mapping relationship with human perception, respectively. Via revealing the potential relationship between natural image and point cloud, we can use the prior knowledge of IQA to guide point cloud distortion evaluation.

 To transfer useful information from the 2D natural image domain to the 3D point cloud domain, the point clouds are first projected into images to satisfy a shared feature extraction and processing network with natural images. Then we design an effective feature generator, i.e., the hierarchical SCNN (H-SCNN), and a sophisticated discriminator, i.e., the conditional-discriminative network, to minimize domain discrepancy and produce robust features. SCNN is a lightweight network proposed in \cite{SCNN} for evaluating natural image quality. The motivation of H-SCNN is that multiscale features are canonical for both natural images and point clouds
in terms of the full-reference IQA and PCQA\cite{wang2003multiscale}\cite{yang2019modeling}\cite{yang2020inferring}. Considering the ultimate purpose of our model is quality assessment rather than simply minimizing the domain discrepancy, we propose a conditional cross entropy loss (CCEL) for the conditional-discriminative network. While reducing the domain discrepancy, we penalize the features that are less irrelevant for quality regression and improve the final results.

\begin{figure*}[pt]
\setlength{\abovecaptionskip}{0.cm}
\setlength{\belowcaptionskip}{-0.cm}
	\centering
		\includegraphics[width=0.8\linewidth]{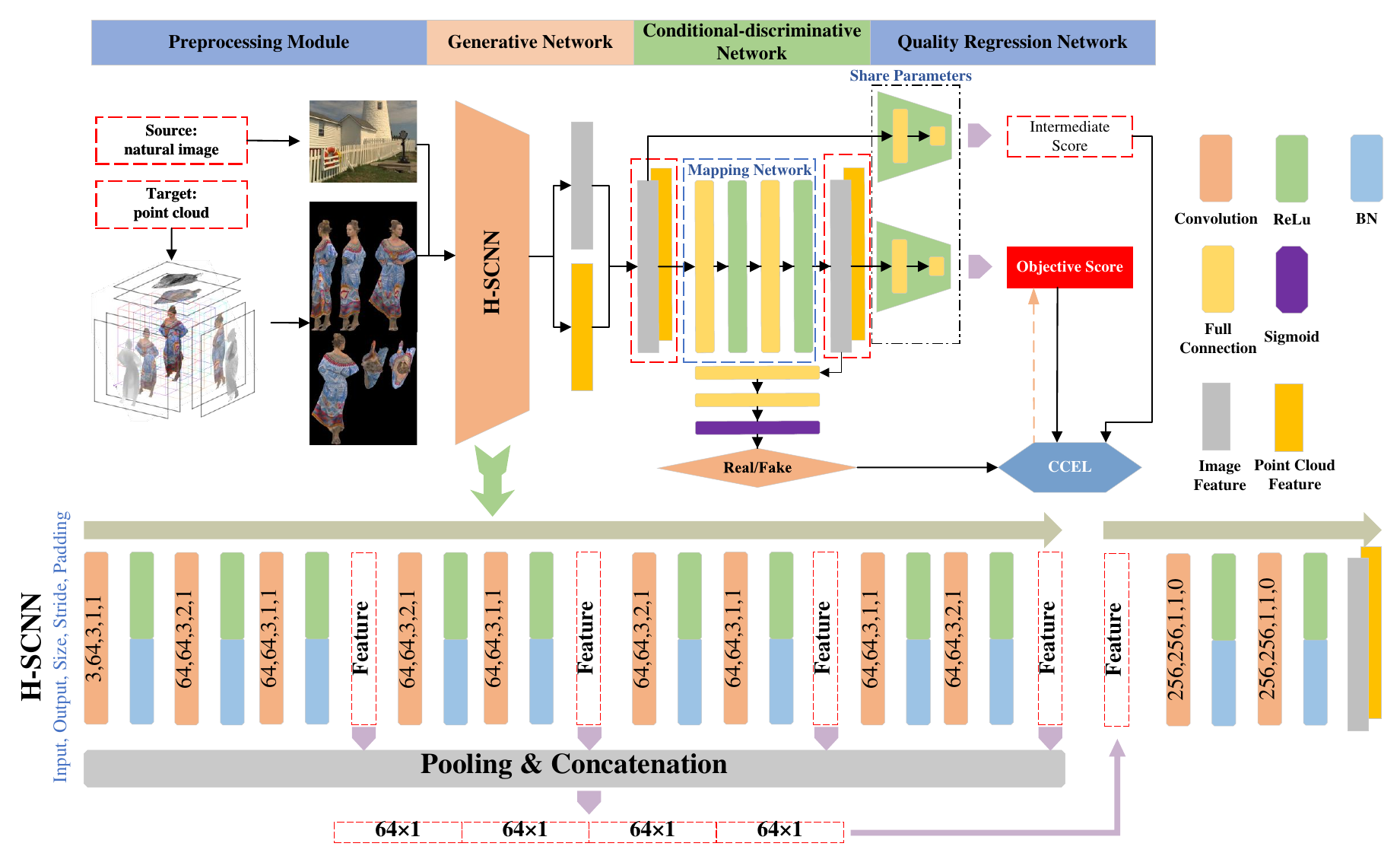}
	\caption{An illustration of our proposed IT-PCQA. It consists of four strongly related substeps. (a) Preprocessing module is used to generate multi-perspective images based on the point cloud sample. (b) A generative network is used to extract hierarchical features from both source and target domains. (c) The conditional-discriminative network is in a position to connect feature generative network and quality regression network, in which the features are refined to match domain shift and liable to regression network produce accurate results. Specifically, the size of FCs in mapping network are both $256\times256$, and the FCs in front of Sigmoid are $256\times64$ and $64\times1$.  (d)  Quality regression network is the endgame of the whole model, which achieves the prediction of quality. Specifically, the two FCs with $256\times128$ and $128\times1$. }
	\label{fig:scheme}
\end{figure*}

In the proposed model called image transferred point cloud quality (IT-PCQA), which is shown in Fig. \ref{fig:scheme}.  we first use  six-perpendicular-projection to generate multi-prospective texture images for point clouds. Then, the images are sent into H-SCNN to generate latent features. Next, the conditional-discriminative network matches the feature distributions of source and target domains and refine the features so that they are more related to objective score regression. Finally, the quality regression network maps the features to objective scores.

The main contributions of this paper include:

$\bullet$ We propose a novel unsupervised no-reference PCQA framework, called IT-PCQA, which uses the ground truth of IQA to emulate the human perception of 3D point clouds. To the best of our knowledge, this is the first unsupervised deep-learning-based no-reference PCQA metric via domain adaption, which reveals the potential relevance between 2D and 3D media in the field of quality assessment.

$\bullet$   To realize transfer learning concerning using the natural images as the source domain, we use multi-perspective images of the point cloud as the network input. To extract effective features for both natural and multi-perspective images, we propose a hierarchical feature encoder, H-SCNN, to fuse multiscale structure features. Compared with classical image feature encoder, e.g., VGG16, ResNet, AlexNet, and original SCNN, the proposed H-SCNN presents obviously better performance on PCQA.

$\bullet$  Considering that regressing objective scores is our final goal, we propose a conditional-discriminative network to produce robust features for quality prediction while minimizing the domain discrepancy.  A novel loss function, i.e., CCEL, is advocated to penalize the samples that negatively impact the perception prediction and stabilize the results of adversarial learning.

$\bullet$ We compare the proposed IT-PCQA with multiple full-reference metrics and two typical no-reference PCQA baselines. Experimental results show that the proposed IT-PCQA presents better performances than the two typical supervised learning networks in most cases, and it is even superior to some full-reference metrics.

\section{Related Works}

\textbf{Point Cloud Quality Assessment}
For PCQA, the full-reference metric is firstly studied due to its urge requirement in point cloud compression.    The classic methods are point-based full-reference metrics proposed by MPEG, such as p2point~\cite{Mekuria2016Evaluation}, p2plane ~\cite{tian2017geometric} and $\rm PSNR_{YUV}$~\cite{torlig2018novel}. Although these metrics have low computation complexity, they present unstable performance when facing multiple types of distortion according to a serial subjective experiment \cite{alexiou2017subjective,alexiou2017performance}. These point-based metrics ignore the fact that our HVS is more sensitive to structure features, only considering single point/pixel features may incur spurious prediction. Inspired by SSIM \cite{wang2004ssim}, a series of PCQA metrics are proposed that taking structural features into consideration and realize more robust performance~\cite{meynet2020pcqm}\cite{yang2020inferring}\cite{yang2020MPED}\cite{Zhang2021MSGRAPHSIM}.

However, the exploration of no-reference PCQA is relatively late due to the limitation of prior knowledge and database.
Therefore, \cite{Chetouani2021nopcqa} split point cloud into multiple local patches and used low-level patch-wise features (e.g., geometric distance, local curvature) to train a convolution neural network (CNN).  \cite{liu2021csvt} proposed to use multiview projection to realize point cloud data argumentation. The above two types of research enrich the database via reproducing original samples, which cannot fundamentally mitigate the intractable problem that lacking sufficient data.
\cite{liu2020LSPCQA} proposed to ameliorate the scarcity of MOSs via pseudo-MOSs.
However, the accuracy of pseudo MOSs is highly depends on the performance of full-reference metrics.

In this work, we propose to use UADA to solve { the problem of insufficient labels.}
Considering the judge of quality assessment is HVS for different forms of media (e.g., image, video, and point cloud), the labels of IQA can reveal the characteristics of HVS and contribute to the quality assessment of point cloud.

\textbf{Domain Adaption}
DA refers to the goal of learning a concept from labeled data in the source domain that performs well on a different but related target domain. DA can reduce the need for costly labeled data in the target domain.  Common practices of DA are matching the feature distribution between the source and a target domain, such as map distributions of two domains via minimizing maximum mean discrepancy (MMD)~\cite{Tzeng2014MMD}, correlation alignment~\cite{Sun2016Corre} and Kullback Leibler divergence~\cite{Zhuang2015kl}. In recent years, adversarial DA has become an increasingly popular incarnation of this type of approach, which minimizes the domain discrepancy through an adversarial objective concerning a domain discriminator ~\cite{Tzeng_2017_CVPR}.

DA has been widely used in image classification and recognition \cite{pmlr-v37-ganin15}\cite{NIPS2016_45fbc6d3}. As we introduced in the section \ref{section:introduction}, point clouds lack sufficient labels while images can provide rich ground truth. With the same purpose under a unified evaluation system, i.e., HVS, using DA is a plausible method to solve the problem of no-reference PCQA. According to the best of our knowledge, the only application of DA on quality assessment is \cite{ChenDA}, in which the authors used MMD as a loss function to predict screen content image quality via treating natural images as source domain. In this paper, we discuss the application of DA between 2D and 3D perception, which is more complex than the case in \cite{ChenDA}. According to the best of our knowledge, there is no study on PCQA via DA. But in \cite{daICCVws}, the authors used the bird's view of a point cloud image to enhance the original LiDAR point cloud data via DA. Besides, \cite{PCRE}\cite{PCREG}  studied 3D reconstruction and recognition based on single or multiview images. Therefore,  using DA to solve the problem of PCQA is feasible.


\section{Problem Formulation}
\label{sec:problem_formulation}



In unsupervised domain adaption, source domain are given as $\Dset\{(\x_i^s,\y_i^s)\}_{i=1}^{n_s}$ and target domain $\Tset\{\x_j^t\}_{j=1}^{n_t}$, in which $\x_i^s$ and $\x_j^t$ represent samples in source and target domain, and $\y_i^s$ denotes the label of $\x_i^s$. Assuming the source and target domain are sampled from the distribution $\Pset(\x^s, \y^s)$ and $\Qset(\x^t, \y^t)$.


For quality assessment, the ultimate goal of DA is to learn a feature extractor $G(\cdot)$ and a regression module $R(\cdot)$ to minimize the expected target risk $\varepsilon_{(\x_i^t,\y_i^t)\sim\Tset}[\Lset_{sim}\{R(G(\x_i^t)),\y_i^t)\}]$, where $\Lset_{sim}\{\cdot\}$ is an index which can evaluate the discrepancy between the subjective and objective scores, such as Pearson linear correlation coefficient (PLCC), Spearman rankorder correlation coefficient (SROCC) and root mean squared error (RMSE). The most classical method to realize above goal is reduce the domain distribution discrepancy by jointly constraining the distance between $\Pset(\x^s)$ and $\Qset(\x^t)$ in the feature space by a MMD loss.


Adversarial domain adaption is an integration of adversarial learning and DA. A domain discriminator $D(\cdot)$ is learned by minimizing the classification error of distinguishing target domain samples from source domains. Through fully confusing the discriminator, we hope the distribution of source and target domain are sufficiently similar. Generally, the cross entropy loss is used to penalize negative discrimination, e.g.,
$$\Lset_{adv} = -\mathbb{E}_{\x_i^s\sim\Dset}log[D(G(\x_i^s))]- \mathbb{E}_{\x_j^t\sim\Tset}log[1-D(G(\x_j^t))].$$

\section{Our Approach}\label{sec:our_approach}
In this section, we introduce the proposed IT-PCQA in detail. The framework of the proposed model is illustrated in Fig. \ref{fig:scheme}, i.e., it has the point clouds preprocessing module, the feature generative network $G$, the conditional-discriminative network $D$, and the quality regression network $R$.

The preprocessing module is used to transfer point clouds to multi-perspective images, and the quality regression network is used to produce the final objective scores based on the generated features. The main task of IT-PCQA is to design an effective feature generator $G$ and a sophisticated discriminator $D$. For the generator $G$, we resort to the achievement of both full-reference IQA and PCQA where multiscale features provide robust performance under manifold distortions. Therefore, we propose a hierarchical feature generator, i.e., H-SCNN, to produce accurate feature representation. For the discriminator $D$, we realize that only minimizing the domain discrepancy is not enough for quality assessment. Therefore, we propose a conditional-discriminative network that uses quality assessment indexes to screen out the features which contribute less to regressing an accurate objective score.


 \subsection{Point Cloud Preprocessing Module}\label{sec:pcpre} 

 To share a common feature encoder, image and point cloud need to be cast into the same data form. Therefore,  we first project the 3D point cloud onto six perpendicular planes of a cube. Then, the six images are spliced together to form a multi-perspective image as the network input, as shown in Fig. \ref{fig:scheme}.

The main problem to answer is whether the projected images can serve as a substitute for original point clouds. The 3D-to-2D projection will unavoidably introduce information loss. Generally, if the information loss does not influence the quality prediction results, we would tend to believe that using projected images as a substitute is reasonable.
In \cite{yang2020predicting}\cite{alexiou2019exploiting}, the authors claim that six projection planes are enough to derive the final results through evaluating several objective metrics based on these images. Although they get comparable performance via projection from six perpendicular planes, the argumentation method may introduces bias.

\begin{figure}[pt]
\setlength{\abovecaptionskip}{0.cm}
\setlength{\belowcaptionskip}{-0.cm}
	\centering
		\includegraphics[width=0.9\linewidth]{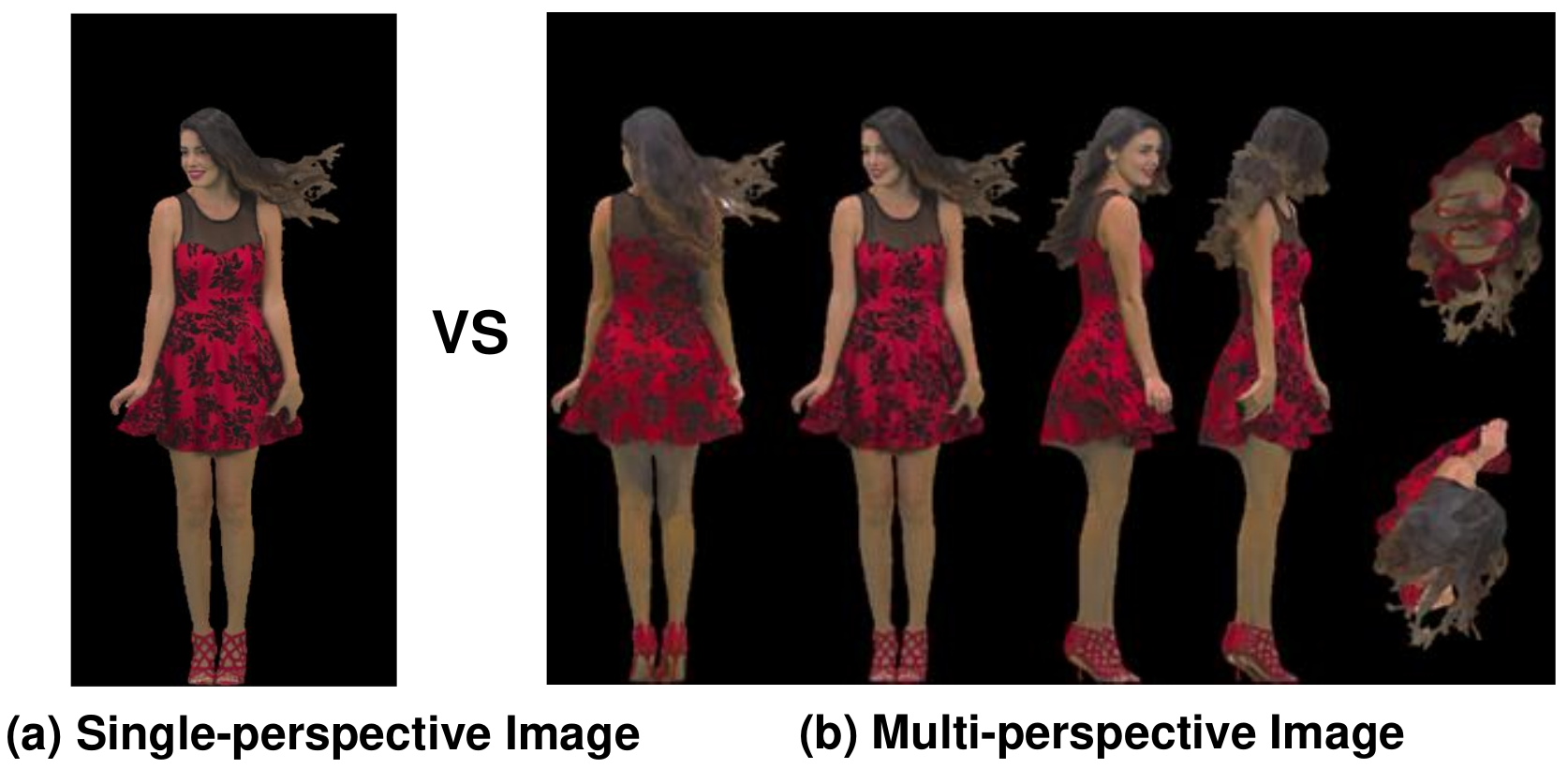}
	\caption{Presentation of projected images from ``RedandBlack''. }
	\label{fig:view}
\end{figure}

Therefore, we conduct a compact subjective experiment to demonstrate the reasonableness of this approach from the basic level. We present point cloud samples in the form of 3D data (e.g., rendered by MeshLab with free-viewpoint) and 2D data (e.g., rendered by image viewer after projection, including single-perspective (2D-1) and multi-perspective images (2D-2)), such as Fig. \ref{fig:view} shows. After pre-training, we ask the subjective experimenter whether they can perceive the samples are distorted, and if yes, what kinds of distortion are injected and the corresponding subjective scores. The scoring method refer to \cite{yang2020predicting}. Note via comparing the statistical data concerning rendered as a 3D point cloud, 2D-1 and 2D-2, we can quantify the influence of projection strategies.

\textbf{Sample Preparation:}
{ To ensure the reliability of the subjective experiment, the reference samples should have different visual characteristics, and the distortion types need to cover the practical cases as much as possible.}
Therefore, we choose three point cloud samples with obviously different characteristics, e.g., ``RedandBlack'', ``Soldier'' and ``Statue '' as references \cite{yang2020predicting}. Then, we choose seven typical distortion types refer to current prevalent databases, including \textit{Octree (OT)}, \textit{DownSampling (DS)}, \textit{Geometrical Gaussian Noise (GN)}, \textit{Color Noise (CN)},  \textit{Color Quantization Noise (QN)}, \textit{Local Loss (LL)} and \textit{V-PCC (VPCC)}. These distortions cover the basic cases of point cloud capture, compression, and reconstruction. Other complex distortions are the superimposed cases of these basic types. For OT, DS, GN, and CN, we use the function scripts provided by \cite{yang2020predicting} and process the reference samples into four different levels. Similarly, QN, VPCC, and LL are generated via the scripts provided by \cite{liu2020LSPCQA}. In total, we prepare $3\times7\times4=84$ distortion samples as test sequences. Based on 3 reference and 84 distorted samples, we use projection method proposed by \cite{yang2020predicting} to generate single- and multi-perspective images. We integrate a visual graphical interface to collect data.

\textbf{Subjective Experiment:}
Considering that experimenters may do not have any prior knowledge about point clouds, we use the sequence ``Soldier'' as a training sequence. Specifically, we use both 3D and 2D data rendering methods to demonstrate the characteristics of the point cloud and its distortion. After training, we use ``RedandBlack'' and ``Statue'' as test sequences. To avoid the cross-influence between different presentation methods, e.g., an experimenter is easier to distinguish the distortion based on the projected image if he/she has viewed the same distorted samples in the form of 3D, we ensure each experimenter only views point clouds in a single presentation method. We collect at least 16 reliable experiment results for each sample in the same presentation to ensure the credibility of experimental data.

\textbf{Result and Conclusion:}
The experiment results are shown in Table \ref{Table:sub}. Acc($\%$) means the accuracy of distortion recognition, and PLCC and SROCC mean the correlation between the subjective score of 2D-2 (2D-1) and that of 3D. We see that: i) 2D-2 presents close performance with 3D on distortion classification, while 2D-1 shows an obvious accuracy decrease; ii) The subjective scores obtained via 2D-2 presentation show a higher correlation with 3D presentation than that obtained via 2D-1 presentation, e.g., 2D-2 provides higher PLCC and SROCC. It means 2D-2 is a better projection strategy that can provide sufficient information for distortion detection and perception quantification as 3D representation.

\begin{table}[h]
\setlength{\abovecaptionskip}{0.cm}
\setlength{\belowcaptionskip}{-0.cm}
	\centering
	\begin{scriptsize}
	\setlength{\tabcolsep}{0.3mm}{
		\begin{tabular}{c|c|c|c|c|c|c}
			\hline
			 & \multicolumn{3}{|c|}{RedandBlack} & \multicolumn{3}{|c}{Statue} \\ \hline
		 &Acc($\%$)&PLCC&SROCC&Acc($\%$)&PLCC&SROCC  \\  \hline
		3D &84&-&-&83&-&-	\\ \hline
		2D-1 &81&0.7017&0.7240&79&0.7371&0.7033\\ \hline
		2D-2&83&0.8611&0.8526&82&0.9140&0.8832\\
\hline
	\end{tabular}}
	\end{scriptsize}
	\caption{Subjective experiment results.
	The 2D-2 presents close performance for distortion recognition accuracy and higher correlation for subjective scores with 3D, which means the multi-perspective images can provide sufficient information as original point clouds.} \label{Table:sub}
\end{table}


\subsection{Feature Generative Network}
Feature generative network $G$ aims to produce robust features for quality regression. Previous researches have demonstrated that deep neural networks present compelling performance on IQA. In this paper, we hope the $G$ can produce effective features for both natural and point clouds projected images.

Refer to the IQA metrics, such as MS-SSIM \cite{wang2003multiscale} and MSEA\cite{yang2019modeling}, multiscale structural features (MSF) show a strong correlation with HVS. Similarly, MS-GraphSIM\cite{Zhang2021MSGRAPHSIM} and MPED \cite{yang2020MPED} indicate that MSF is also capable of PCQA.

Therefore, we propose a hierarchical SCNN (H-SCNN) network as a feature encoder. SCNN is a lighter network proposed in \cite{SCNN}, which offers pronounced performance on both synthetically and authentically distorted images.
Using the prior knowledge of SCNN as a reference, we incorpoarte hierarchical feature fusion and introduce the H-SCNN as Fig. \ref{fig:scheme} shows. We introduce 9 convolution layers, each convolution layer is followed by batch normalization (BN) and rectified linear unit (ReLU) operations. The convolution results can be regarded as a sketching feature map \cite{sketchingSPM} of the input image, e.g., $\C_{j}(\x)$ and $j$ represents the number of layers of convolution. The sketch of the images is a dimensionality-reduced linear projection of the multiple sketching feature maps. Specifically, We extract the 3rd, 5th, 7th, and 9th layered features, and after average pooling, we obtain four latent features with dimensions of $64\times1$. Then, we concatenate four features directly, and twofold convolution layers are followed to generate the final encoded features,
\begin{align}\label{eq:sketch}
\setlength{\abovecaptionskip}{0.cm}
\setlength{\belowcaptionskip}{-0.cm}
G(\x) =\Gamma\{ \Phi_{3}(\x)\oplus \Phi_{5}(\x)\oplus \Phi_{7}(\x)\oplus \Phi_{9}(\x)\},
\end{align}
in which $\Phi_{j}=Ave\{\C_{j}(\x)\}$, $Ave\{\cdot\}$ represents average pooling, $\oplus$ represents concatenation and $\Gamma\{\cdot\}$ represents the final twofold convolutions.


Compared with the original SCNN which only uses a single scale feature, the proposed H-SCNN fuses hierarchical features to produce a judiciously latent feature representation for images. In section \ref{sec:abla}, we illustrate the superiority of the proposed H-SCNN.

\subsection{Conditional-discriminative Network}\label{sec:conditional}

To minimize the domain discrepancy across source and target domains and produce features valuable for score regression, we design a discriminator $D$, i.e., the conditional-discriminative network, in this part. We inject task-specific conditional factors into adversarial learning to stabilize the results of objective score regression.

 Essentially, the discriminator needs to be fully 'misled' so that it cannot distinguish whether the generated features are from a real or fake sample. In our work, we regard the samples from the source domain, i.e., images, are real and these of the target domain, i.e., point clouds, are fake. Generally, this is the generalized form of the utilization of adversarial learning in the classification task. However, quality assessment aims to predict the perceptual result for a media content, which is a degree of the description of the quality rather than simply a ``Yes'' or ``No''. The encoded features are not only needed to fulfill domain matching but also required to facilitate quality score regression.

Therefore, we use the quality assessment evaluation indexes, such as SROCC, as indicators to help train the discriminator and propose a novel loss function, i.e., CCEL, based on the conventional cross entropy loss. While reducing the domain discrepancy, we attempt to boost the performance of $G$ and the features refined by $D$ on score prediction.


Specifically, the output of the feature generative network, i.e., $G(\x)$, is first fed into a feature mapping network $M$, which consists of a twofold full connection  - ReLU layers. The mapped features are kept dimension-unchanged after $M$ (blue dotted box in Fig. \ref{fig:scheme}). As we mentioned before, to make sure the generated features are conducive to score regression, we penalize the features which are impotent to prediction accuracy. This could be formulated as
\begin{align}\label{eq:CCEL}
\Lset_{CCEL} = &-\mathbb{E}_{\x_i^s\sim\Dset}log[|D(M(G(\x_i^s)))-d|]\\
&- \mathbb{E}_{\x_j^t\sim\Tset}log[1-D(M(G(\x_j^t)))].\nonumber
\end{align}
$d$ is a flag that indicates whether the features are promoted after mapping with the definition
\begin{equation}\label{eq:d}
d=
\begin{cases}
1& \text{if } {\Lset_{sim}}\{R(M(G(\x^s))),\y^s\} > \\ &{\Lset_{sim}}\{R(G(\x^s)),\y^s\}+\epsilon\\
0& \text{otherwise.}
\end{cases}
\end{equation}
$R$ is the quality regression network which will be introduced in section \ref{sec:r}. As introduced in section \ref{sec:problem_formulation}, $\Lset_{sim}\{\cdot\}$ can be instantiated as PLCC, SROCC, or RMSE. $\epsilon$ is a threshold that controls the scale of punishment. The general idea behind this formulation is that we leverage the joint variable of feature regression results $R(G(\x^s))$ and $R(M(G(\x^s)))$, and classifier prediction $D(M(G(\x_j^t)))$. If the regression results based on $R(M(G(\x^s)))$ present better performance than $R(G(\x^s))$, it means the feature mapping network $M$ show positive impact on $R$, and the $G(\x^s)$ are relatively weak. We regard the cases that  $G(\x^s)$ are relatively weak as negative samples, otherwise as positive samples. We use Eq. \eqref{eq:d} to penalize the negative samples (i.e., $d=0$) and the opposite of positive samples (i.e., $d=1$). Therefore, this conditional loss function can make sure the adversarial learning process is definitely beneficial to the quality regression, which could also be regarded as a relaxation strategy to stabilize the quality-related discrimination.

\subsection{Quality Regression Network}\label{sec:r}
The quality regression network $R$ is more compact than other parts of the proposed IT-PCQA. We use twofold full connection layers to regress an objective score from a latent feature generated jointly by feature generative network $G$ and conditional-discriminative network $D$.

Given a set of distorted images from target source, e.g., $\{\x^s_i, \y_i\}^{n_s}_{i=1}$, we can obtain a latent feature $\I$ for each image via previous networks, e.g., $\{\x^s_i|\I_i, \y_i\}^{n_s}_{i=1}$, $\I_i=M(G(\x_i^s))$. We train the regression network via
\begin{equation}\label{eq:r}
\setlength{\abovecaptionskip}{0.cm}
\setlength{\belowcaptionskip}{-0.cm}
\Lset_{R}={\rm min} [\frac{1}{n_s}\sum_{i}^{n_s}(R(\I_i)-\y_i)^2].
\end{equation}

\section{Experiment}\label{sec:exp}
\subsection{Experiment Setups}
\subsubsection{Databases}\label{sec:database}

To test the proposed method, we use four independent databases, including two for natural images (TID2013\cite{tid2013} and LIVE\cite{LIVE}) and two for point clouds (SJTU-PCQA\cite{yang2020predicting} and WPC\cite{liu2021csvt}).

\textbf{TID2013.} The TID2013 database consists of 25 reference images. Each image is processed with twenty-four different types of distortion under five different levels. In all, there are 3000 distorted images.

\textbf{LIVE.} The LIVE includes 29 reference images, and five different distortions are injected: JPEG, JPEG2000, additive white Gaussian noise, Gaussian blur, and Rayleigh fast-fading channel distortion. In all, there are 982 distorted natural images.

\textbf{SJTU-PCQA.} The SJTU-PCQA includes 8 reference point clouds. Each native point cloud sample is augmented with seven different types of impairments under six levels, including four individual distortions
and three superimposed distortions.
In all, there are 336 distorted point cloud samples.


\textbf{WPC} There are 20 reference point cloud samples in WPC, and four distortion types are considered, including Downsampling (60 degraded samples), Gaussian white noise (180 degraded samples), G-PCC (240 degraded samples), and V-PCC (180 degraded samples). In all, there are 660 distorted samples.

\subsubsection{Implementation Details}
We implement our model by PyTorch. The natural and projected images are resized to $224\times224\times3$ as the input of H-SCNN. The batch size is set as 16. The overall loss function of all training samples is given by $\Lset_{all}=\mu_1\Lset_{CCEL}+\mu_2\Lset_{R}$, $\mu_1=\mu_2=1$. We set $\Lset_{sim}\{\cdot\}$ in Eq. \eqref{eq:d} as SROCC, and $\epsilon=0.1$.

To compare the proposed IT-PCQA with other no-reference metrics which based on training with labeled data, we split the two databases of point cloud with $3:1$, e.g., $75\%$ for training and $25\%$ for testing (Note for the proposed IT-PCQA, we do not need to split database, the split just for performance compare).


\subsection{Prediction performance}
We compare our IT-PCQA with another 9 state-of-the-art distortion quantifications, 7 full-reference metrics, including MSE-P2point (M-p2po), MSE-P2plane (M-p2pl), Hausdorff-P2point (H-p2po), Hausdorff-P2plane (H-p2pl), $\mathrm{PSNR_{YUV}}$, PCQM, GraphSIM , and two no-reference metrics \cite{liu2020LSPCQA}, e.g., PointNet-based (PointNet-PCQA) and SparseNet-based (SparseNet-PCQA) metrics. Note these two metrics need the assistance of label.

\begin{table}[h]
\setlength{\abovecaptionskip}{0.cm}
\setlength{\belowcaptionskip}{-0.cm}
	\centering
	\begin{scriptsize}
	\setlength{\tabcolsep}{0.7mm}{
		\begin{tabular}{c|c|c|cc}
			\hline
			\multicolumn{3}{c|}{Target domain: SJTU-PCQA} & {PLCC} &{SROCC}  \\ \hline
			\multirow{7}{*}{{Full-reference}}  & \multicolumn{2}{|c|}{M-p2po}  &0.88&0.78\\
			& \multicolumn{2}{|c|}{M-p2pl}  &0.77&0.63  \\ 
			& \multicolumn{2}{|c|}{H-p2po}  &0.60&0.58  \\
		& \multicolumn{2}{|c|}{H-p2pl}  &0.65&0.62  \\ 
			& \multicolumn{2}{|c|}{$\mathrm{PSNR_{YUV}}$}   &0.65&0.62 \\
			& \multicolumn{2}{|c|}{PCQM}   &0.84&0.83\\
			& \multicolumn{2}{|c|}{GraphSIM}   &\textbf{0.91}&\textbf{0.89}  \\ \hline
			
			\multirow{4}{*}{{No-reference}} &  \multicolumn{2}{|c|}{PointNet-PCQA}  &0.46&0.44\\ 
			&\multicolumn{2}{|c|}{SparseNet-PCQA}  &0.47&0.44\\ 
			& \multicolumn{2}{|c|}{IT-PCQA (Training on TID2013)} &\textbf{0.58}&\textbf{0.63} \\
			& \multicolumn{2}{|c|}{IT-PCQA (Training on LIVE)} &0.57&{0.51} \\
\hline
	\end{tabular}}
	\caption{Model performance for point clouds samples in SJTU-PCQA database. The Proposed IT-PCQA presents comparable performance with partial full-reference metrics, and best performance on no-reference metrics. } \label{Table:SJTU}
	\end{scriptsize}
\end{table}


We treat SJTU-PCQA and WPC as target domains, and use TID2013 and LIVE as the source domain, respectively. The results are shown in Table \ref{Table:SJTU} and \ref{Table:WPC}. We see that: i) Overall, full-reference metrics present better performance than no-reference metrics; ii) IT-PCQA (Training on TID2013) presents better PLCC and SROCC than IT-PCQA (Training on LIVE) in average. For example, (0.58, 0.63) vs (0.57, 0.51) on SJTU-PCQA, and (0.54, 0.54) vs (0.31, 0.28) on WPC. Refer to section \ref{sec:database}, TID2013 provides more types of distortion and more distorted samples than LIVE. Therefore, the richness of the source domain can impact the final results for the quality assessment task; iii) The performance of IT-PCQA (Training on TID2013) is better than some full-reference metrics, such as H-p2po, H-p2pl, and $\rm PSNR_{YUV}$ on WPC;
iv) Besides, we give some illustrative examples of distorted samples of SJTU-PCQA in Fig. \ref{fig:mos}. Corresponding Moss and IT-PCQA values are provided. The proposed IT-PCQA presents highly consistent with the MOSs.
			

\begin{table}[h]
\setlength{\abovecaptionskip}{0.cm}
\setlength{\belowcaptionskip}{-0.cm}
	\centering
	\begin{scriptsize}
	\setlength{\tabcolsep}{0.7mm}{
		\begin{tabular}{c|c|c|cc}
			\hline
			\multicolumn{3}{c|}{Target domain: WPC} & {PLCC} &{SROCC}  \\ \hline
			\multirow{7}{*}{{Full-reference}} & \multicolumn{2}{|c|}{M-p2po} &0.61&0.58\\
			& \multicolumn{2}{|c|}{M-p2pl}  &0.63&0.59  \\ 
			& \multicolumn{2}{|c|}{H-p2po}  &0.51&0.46  \\
			& \multicolumn{2}{|c|}{H-p2pl} &0.55&0.48  \\ 
			& \multicolumn{2}{|c|}{$\mathrm{PSNR_{YUV}}$}   &0.46&0.47 \\
			& \multicolumn{2}{|c|}{PCQM}   &\textbf{0.74}&\textbf{0.75} \\
			& \multicolumn{2}{|c|}{GraphSIM}   &\textbf{0.74}&\textbf{0.75}  \\ \hline
			
			\multirow{3}{*}{{No-reference}} & \multicolumn{2}{|c|}{PointNet-PCQA}  &0.23&0.28\\ 
	& \multicolumn{2}{|c|}{SparseNet-PCQA}  &0.16&0.24\\ 
			& \multicolumn{2}{|c|}{IT-PCQA (Training on TID2013)} &\textbf{0.55}&\textbf{0.54} \\
			& \multicolumn{2}{|c|}{IT-PCQA (Training on LIVE)} &0.31&0.28 \\
\hline
	\end{tabular}}
	\caption{Model performance for point clouds samples in WPC database. The proposed IT-PCQA presents obvious superiority. } \label{Table:WPC}
	\end{scriptsize}
\end{table}

In a summary, the proposed IT-PCQA presents robust and competitive performance on two independent databases compared over other no-reference metrics and even full reference models. It validates our motivation that there is definitely a connection between 2D and 3D perception, and we can infer point cloud quality via utilizing the fruitful achievement on IQA.

\begin{figure}[pt]
\setlength{\abovecaptionskip}{0.cm}
\setlength{\belowcaptionskip}{-0.cm}
	\centering
		\includegraphics[width=0.9\linewidth]{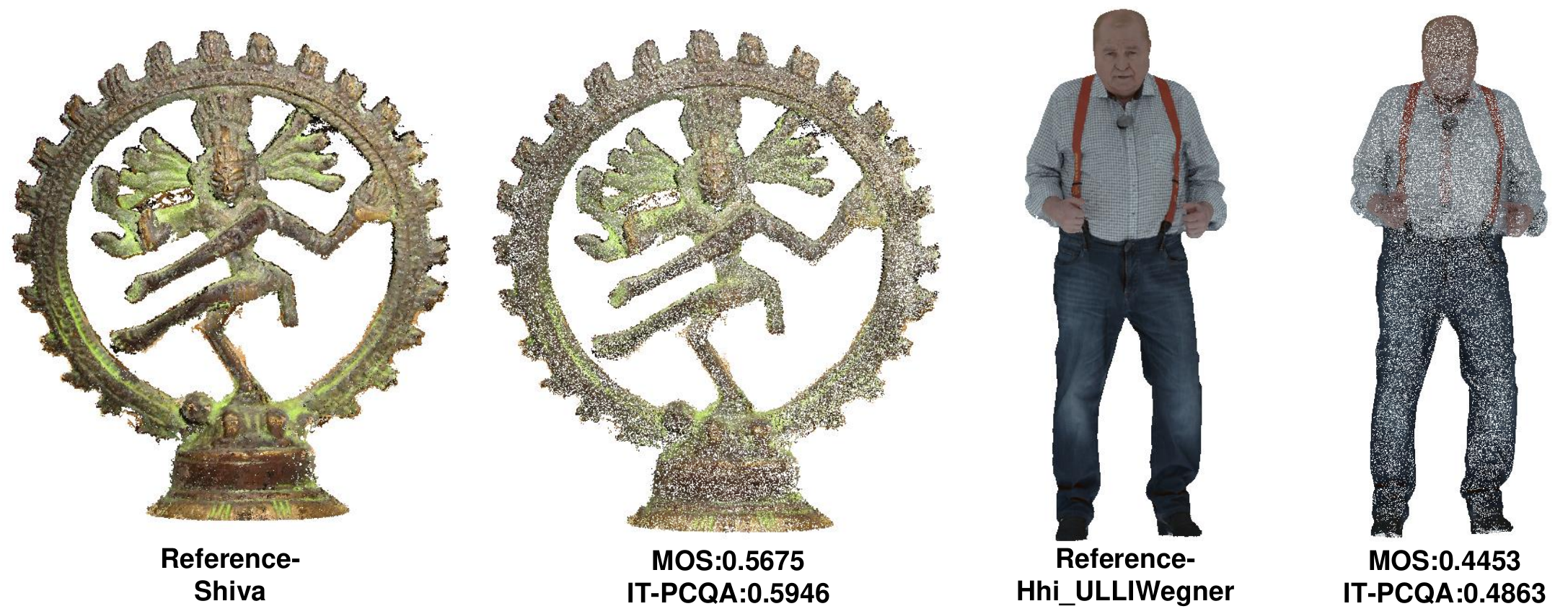}
	\caption{Exemplified point cloud samples in SJTU-PCQA. The associated MOS and IT-PCQA values are highly consistent.  }
	\label{fig:mos}
\end{figure}

\subsection{Ablation Study}\label{sec:abla}
This section examines the proposed IT-PCQA by dissecting and reassembling its modules to demonstrate its generalization and efficiency on human perception prediction.

\subsubsection{Hierarchical Feature Encoder}
In this section, we demonstrate the superiority of the proposed H-SCNN.

We use five different feature encoders as generators, including SCNN\cite{SCNN}, VGG16\cite{vgg16}, ResNet50\cite{resNet}, AlexNet\cite{AlexNet} and the proposed H-SCNN. We use TID2013 as source domain and WPC as target domain.
 The results are shown in Table \ref{tab:ab_hi1}. We see that the proposed hierarchical feature presents obvious superiority compared with other feature encoders. For example, the proposed H-SCNN presents the best PLCC and SROCC, i.e., (0.5491, 0.5371), while AlexNet and VGG16 are only (0.1407, 0.2107) and (0.1448, 0.2511).

 \begin{table}[t]
 \setlength{\abovecaptionskip}{0.cm}
\setlength{\belowcaptionskip}{-0.cm}
	\centering
	\begin{tabular}{c|cc}
		\hline
		Feature Generator & PLCC & SROCC \\ \hline
 		VGG16 & 0.1448 & 0.2511 \\ \hline
		ResNet50 & 0.4266 & 0.4360  \\ \hline
		AlexNet & 0.1407 & 0.2107 \\ \hline
				SCNN & 0.4270 & 0.4276  \\ \hline
		H-SCNN & {\textbf{0.5491}} & {\textbf{0.5371}}  \\ \hline
	\end{tabular}
	\caption{Model performance with different feature generators on WPC. The proposed H-SCNN presents best performance.} \label{tab:ab_hi1}
\end{table}

\subsubsection{The Amount of Projection Planes}

In section \ref{sec:pcpre}, we use subjective experiment to validate the proposed method. In this part, we use objective experiment results to prove the rationality of the proposed multi-perspective images (i.e., 2D-2). Specifically, we use the single perspective images (i.e., 2D-1) as input to repeat the experiment and the results are shown in Table \ref{Table:ab_hi}.
\begin{table}[h]
\setlength{\abovecaptionskip}{0.cm}
\setlength{\belowcaptionskip}{-0.cm}
	\centering
	\begin{scriptsize}
	\setlength{\tabcolsep}{1mm}{
		\begin{tabular}{c|cc|cc}
			\hline
			\multicolumn{3}{c|}{TID2013 to SJTU-PCQA} & \multicolumn{2}{|c}{LIVE to SJTU-PCQA}  \\ \hline
			&PLCC&SROCC&PLCC&SROCC\\ \hline
		    2D-1&0.4251&0.4395&0.4018&0.4975\\ \hline
		    2D-2&{\textbf{0.5791}}&{\textbf{0.6342}}&{\textbf{0.5662}}&{\textbf{0.5124}} \\ \hline \hline


		    \multicolumn{3}{c|}{TID2013 to WPC} & \multicolumn{2}{|c}{LIVE to WPC}  \\ \hline
			&PLCC&SROCC&PLCC&SROCC\\ \hline
		    2D-1&0.3841&0.3620&{\textbf{0.3512}}&{\textbf{0.4284}}\\ \hline
		    2D-2&{\textbf{0.5491}}&{\textbf{0.5371}}&0.3099&0.2833 \\ \hline \hline
		
	\end{tabular}}
	\caption{Model performance for different feature types. The proposed hierarchical feature present best overall performance.} \label{Table:ab_hi}
	\end{scriptsize}
\end{table}
We see that the multi-perspective images present better performance than single perspective images in most cases. Single perspective images cannot provide sufficient information to reveal distortion. For example, the distortion type LL might can not be perceived from one certain perspective, while any local loss can be detected via jointly observing six different views.

\subsubsection {Conditional Adversarial Learning}
In this part, we illustrate the effectiveness of the proposed CCEL. Specifically, treating TID3013 as the source domain and WPC as the target domain, we use $\Lset_{R}$ to train the network as a benchmark. In order to better demonstrate the influence of $d$, we use $\Lset_{T_1}=\Lset_{R}+\Lset_{MMD}$ and $\Lset_{T_2}=\Lset_{R}+\Lset_{adv}$ as loss function to repeated the trials. Specifically, we use the output of H-SCNN to calculate the MMD.  Besides, in section \ref{sec:exp}, we set $\Lset_{sim}\{\cdot\}$ as SROCC. To rule out the influence of SROCC itself, we also present the performance of using $\Lset_{T_3}=\Lset_{R}+ \rm SROCC\{R(\I), \y\}$ as loss function.
The results are shown in Table \ref{tab:ab_loss}. We see that: i) $\Lset_{R}$ is a necessary term to guide score regression. However, refer to the performance of $\Lset_{T_1}$ and $\Lset_{T_2}$, simply learning invariant feature representations in the shallow regime can not improve the final performance, even degrade the results; ii) Based on the performance of $\Lset_{T_3}$, we know that point clouds and images exist obviously different characteristics, without effective transfer learning, the prior knowledge on IQA cannot cover the intractable problem on point clouds; iii) Whereby the injecting of conditional adversarial learning, the discriminator is fully confused and the feature generator can produce robust features for quality prediction. Compared with the traditional cross entropy loss, the proposed CCEL injects task-specific conditional factors and improves the overall performance for specific task. As we explain in section \ref{sec:conditional}, the ultimate goal of quality assessment is regressing quality scores. The encoded features not only need to fool the discriminator and match domain distribution but also need to contribute to the perception prediction.

\begin{table}[t]
\setlength{\abovecaptionskip}{0.cm}
\setlength{\belowcaptionskip}{-0.cm}
	\centering
	\begin{tabular}{c|cc}
		\hline
		Loss & PLCC & SROCC \\ \hline
		$\Lset_{R}$ & {{0.3870}} & 0.3657  \\ \hline
 		$\Lset_{T_1}$ & 0.4276 & 0.3838 \\ \hline
		$\Lset_{T_2}$ & 0.3125 & 0.3392  \\ \hline
		$\Lset_{T_3}$ & {{0.0272}} & 0.2233 \\ \hline
		$\Lset_{all}$ & {\textbf{0.5491}} & {\textbf{0.5371}}  \\ \hline
	\end{tabular}
	\caption{Model performance with different loss function on WPC. The proposed CCEL can improve the overall performance significantly.} \label{tab:ab_loss}
\end{table}


\section{Conclusions}\label{sec:con}
In this paper, we propose to use the unsupervised adversarial domain adaptation to solve the problem of no-reference point cloud quality assessment. The proposed model, called IT-PCQA, utilizes the rich prior knowledge in natural images and builds a bridge between 2D and 3D perception in the field of quality assessment. Without the assistance of a large-scale labeled point cloud subjective database, we establish a no-reference PCQA metric whereby transfer learning, and realize accurate quality prediction on multiple independent PCQA databases. Our method reveals the potential connection between different types of media content in the field of quality assessment, and meet the grand challenges faced by no-reference quality metric in a different real-world application.
\section{Acknowledgment}
This paper is supported in part by National Key R\&D Program of China (2018YFE0206700,2018YFB1802201), National Natural Science Foundation of China (61971282, U20A20185), and Scientific Research Plan of the Science and Technology Commission of Shanghai Municipality (18511105402). Corresponding author is Yiling Xu.

{\small
\bibliographystyle{ieee_fullname}
\bibliography{egbib}

\begin{thebibliography}{10}\itemsep=-1pt

\bibitem{alexiou2017subjective}
Evangelos Alexiou.
\newblock On subjective and objective quality evaluation of point cloud
  geometry.
\newblock In {\em Int. Conf. Quality of Multimedia Experience (QoMEX'17)},
  pages 1--3, 2017.

\bibitem{alexiou2019exploiting}
Evangelos Alexiou.
\newblock Exploiting user interactivity in quality assessment of point cloud
  imaging.
\newblock In {\em Int. Conf. Quality of Multimedia Experience (QoMEX'19)},
  pages 1--6, 2019.

\bibitem{alexiou2017performance}
Evangelos Alexiou and Touradj Ebrahimi.
\newblock On the performance of metrics to predict quality in point cloud
  representations.
\newblock In {\em Applications of Digital Image Processing XL}, volume 10396,
  page 103961H, 2017.

\bibitem{Sun2016Corre}
Sun Baochen and Saenko Kate.
\newblock Deep coral: Correlation alignment for deep domain adaptation.
\newblock In {\em European Conf. Computer Vision (ECCV'16)}, pages 443--450.
  Springer, 2016.

\bibitem{NIPS2016_45fbc6d3}
Konstantinos Bousmalis, George Trigeorgis, Nathan Silberman, Dilip Krishnan,
  and Dumitru Erhan.
\newblock Domain separation networks.
\newblock In {\em Advances in Neural Information Processing Systems
  (NeurIPS'16)}, volume~29. Curran Associates, Inc., 2016.

\bibitem{ChenDA}
Baoliang Chen, Haoliang Li, Hongfei Fan, and Shiqi Wang.
\newblock No-reference screen content image quality assessment with
  unsupervised domain adaptation.
\newblock {\em IEEE Trans. Image Processing}, 30:5463--5476, 2021.

\bibitem{ChenLFGW:20}
Siheng Chen, Baoan Liu, Chen Feng, Carlos Vallespi{-}Gonzalez, and Carl
  Wellington.
\newblock 3d point cloud processing and learning for autonomous driving.
\newblock {\em IEEE Signal Processing Magazine}, 2020.

\bibitem{Chetouani2021nopcqa}
Aladine Chetouani, Maurice Quach, Giuseppe Valenzise, and Frédéric Dufaux.
\newblock Deep learning-based quality assessment of 3d point clouds without
  reference.
\newblock In {\em 2021 IEEE Int. Conf. Multimedia Expo Workshops (ICMEW'21)},
  pages 1--6, 2021.

\bibitem{Tzeng2014MMD}
Tzeng Eric, Hoffman Judy, Zhang Ning, Saenko Kate, and Darrell Trevor.
\newblock Deep domain confusion:maximizing for domain invariance.
\newblock {\em arXiv preprint arXiv:1412.3474}, 2014.

\bibitem{Zhuang2015kl}
Zhuang Fuzhen, Cheng Xiaohu, Luo Ping, Pan Sinno~Jialin, and He Qing.
\newblock Supervised representation learning: Transfer learning with deep
  autoencoders.
\newblock In {\em Int. Joint Conf. Artificial Intelligence (IJCAI'15)}, 2015.

\bibitem{pmlr-v37-ganin15}
Yaroslav Ganin and Victor Lempitsky.
\newblock Unsupervised domain adaptation by backpropagation.
\newblock In {\em Proceedings of the Int. Conf. Machine Learning (ICML'15)},
  pages 1180--1189. PMLR, 2015.

\bibitem{sketchingSPM}
Remi Gribonval, Antoine Chatalic, Nicolas Keriven, Vincent Schellekens, Laurent
  Jacques, and Philip Schniter.
\newblock Sketching data sets for large-scale learning: Keeping only what you
  need.
\newblock {\em IEEE Signal Processing Magazine}, 38(5):12--36, 2021.

\bibitem{resNet}
Kaiming He, Xiangyu Zhang, Shaoqing Ren, and Jian Sun.
\newblock Deep residual learning for image recognition.
\newblock In {\em Proceedings of the IEEE/CVF Conf. Computer Vision and Pattern
  Recognition (CVPR'16)}, pages 770--778, 2016.

\bibitem{AlexNet}
Alex Krizhevsky, Ilya Sutskever, and Geoffrey~E Hinton.
\newblock Imagenet classification with deep convolutional neural networks.
\newblock {\em Advances in Neural Information Processing Systems (NeurIPS'12)},
  25:1097--1105, 2012.

\bibitem{scstmm}
Qiaohong Li, Weisi Lin, Jingtao Xu, and Yuming Fang.
\newblock Blind image quality assessment using statistical structural and
  luminance features.
\newblock {\em IEEE Trans. Multimedia}, 18(12):2457--2469, 2016.

\bibitem{lim2020Augmented}
S. {Lim}, M. {Shin}, and J. {Paik}.
\newblock Point cloud generation using deep local features for augmented and
  mixed reality contents.
\newblock In {\em 2020 IEEE Int. Conf. Consumer Electronics (ICCE'20)}, pages
  1--3, 2020.

\bibitem{LQrate}
Qi Liu, Hui Yuan, Raouf Hamzaoui, Honglei Su, Junhui Hou, and Huan Yang.
\newblock Reduced reference perceptual quality model with application to rate
  control for video-based point cloud compression.
\newblock {\em IEEE Trans. Image Processing}, 30:6623--6636, 2021.

\bibitem{liu2021csvt}
Qi Liu, Hui Yuan, Honglei Su, Hao Liu, Yu Wang, Huan Yang, and Junhui Hou.
\newblock Pqa-net: Deep no reference point cloud quality assessment via
  multi-view projection.
\newblock {\em IEEE Trans. Circuits and Systems for Video Technology}, pages
  1--1, 2021.

\bibitem{liu2020LSPCQA}
Y. {Liu}, Q. {Yang}, and Y. {Xu}.
\newblock Point cloud quality assessment: Large-scale dataset construction and
  learning-based no-reference approach.
\newblock {\em arXiv preprint arXiv:2012.11895}, 2020.

\bibitem{2D3Dre}
Priyanka Mandikal, Navaneet K~L, and R. Venkatesh~Babu.
\newblock 3d-psrnet: Part segmented 3d point cloud reconstruction from a single
  image.
\newblock In {\em Proceedings of the European Conf. Computer Vision Workshops
  (ECCVW'18)}, September 2018.

\bibitem{PCRE}
Priyanka Mandikal and Venkatesh~Babu Radhakrishnan.
\newblock Dense 3d point cloud reconstruction using a deep pyramid network.
\newblock In {\em 2019 IEEE Winter Conf. Applications of Computer Vision
  (WACV'19)}, pages 1052--1060, 2019.

\bibitem{2D3Dtracking}
E. Marchand, P. Bouthemy, F. Chaumette, and V. Moreau.
\newblock Robust real-time visual tracking using a 2d-3d model-based approach.
\newblock In {\em Proceedings of the Int. Conf. Computer Vision (ICCV'99)},
  volume~1, pages 262--268 vol.1, 1999.

\bibitem{Mekuria2016Evaluation}
R. Mekuria, Z. Li, C. Tulvan, and P. Chou.
\newblock Evaluation criteria for point cloud compression.
\newblock {\em ISO/IEC MPEG n16332, Geneva, Switzerland}, Feb, 2016.

\bibitem{meynet2020pcqm}
Gabriel Meynet, Yana Nehm{\'e}, Julie Digne, and Guillaume Lavou{\'e}.
\newblock Pcqm: A full-reference quality metric for colored 3d point clouds.
\newblock In {\em Int. Conf. Quality of Multimedia Experience (QoMEX'20)},
  2020.

\bibitem{tid2013}
Nikolay Ponomarenko, Lina Jin, Oleg Ieremeiev, Vladimir Lukin, Karen
  Egiazarian, Jaakko Astola, Benoit Vozel, Kacem Chehdi, Marco Carli, Federica
  Battisti, and C.-C. {Jay Kuo}.
\newblock Image database tid2013: Peculiarities, results and perspectives.
\newblock {\em Signal Processing: Image Communication}, 30:57--77, 2015.

\bibitem{Rusu2011PCL}
R.~B. {Rusu} and S. {Cousins}.
\newblock 3d is here: Point cloud library (pcl).
\newblock In {\em 2011 IEEE Int. Conf. Robotics and Automation (ICRA'11)},
  pages 1--4, 2011.

\bibitem{daICCVws}
Khaled Saleh, Ahmed Abobakr, Mohammed Attia, Julie Iskander, Darius Nahavandi,
  Mohammed Hossny, and Saeid Nahvandi.
\newblock Domain adaptation for vehicle detection from bird's eye view lidar
  point cloud data.
\newblock In {\em Proceedings of the IEEE/CVF Int. Conf. Computer Vision
  Workshops (ICCVW'19)}, Oct 2019.

\bibitem{LIVE}
Hamid~R Sheikh.
\newblock Image and video quality assessment research at live, 2003.
\newblock \url{http://live.ece.utexas.edu/research/quality}.

\bibitem{vgg16}
Karen Simonyan and Andrew Zisserman.
\newblock Very deep convolutional networks for large-scale image recognition.
\newblock {\em arXiv preprint arXiv:1409.1556}, 2014.

\bibitem{tian2017geometric}
Dong Tian, Hideaki Ochimizu, Chen Feng, Robert Cohen, and Anthony Vetro.
\newblock Geometric distortion metrics for point cloud compression.
\newblock In {\em Int. Conf. Image Processing (ICIP'17)}, pages 3460--3464,
  2017.

\bibitem{torlig2018novel}
Eric~M Torlig, Evangelos Alexiou, Tiago~A Fonseca, Ricardo~L de Queiroz, and
  Touradj Ebrahimi.
\newblock A novel methodology for quality assessment of voxelized point clouds.
\newblock In {\em Applications of Digital Image Processing XLI}, volume 10752,
  page 107520I. Int. Soc. Optics and Photonics, 2018.

\bibitem{Tzeng_2017_CVPR}
Eric Tzeng, Judy Hoffman, Kate Saenko, and Trevor Darrell.
\newblock Adversarial discriminative domain adaptation.
\newblock In {\em Proceedings of the IEEE Conf. Computer Vision and Pattern
  Recognition (CVPR'17)}, July 2017.

\bibitem{wang2004ssim}
Zhou Wang, Alan~C Bovik, Hamid~R Sheikh, and Eero~P Simoncelli.
\newblock Image quality assessment: from error visibility to structural
  similarity.
\newblock {\em IEEE Trans. Image Processing}, 13(4):600--612, 2004.

\bibitem{wang2003multiscale}
Zhou Wang, Eero~P Simoncelli, and Alan~C Bovik.
\newblock Multiscale structural similarity for image quality assessment.
\newblock In {\em Asilomar Conf. Signals, Systems \& Computers (ACSS'03)},
  volume~2, pages 1398--1402, 2003.

\bibitem{DomainSurvey}
Garrett Wilson and Diane~J. Cook.
\newblock A survey of unsupervised deep domain adaptation.
\newblock {\em ACM Trans. Intelligent Systems and Technology}, 11(5), July
  2020.

\bibitem{yang2020predicting}
Qi Yang, Hao Chen, Zhan Ma, Yiling Xu, Rongjun Tang, and Jun Sun.
\newblock Predicting the perceptual quality of point cloud: A 3d-to-2d
  projection-based exploration.
\newblock {\em IEEE Trans. Multimedia}, pages 1--1, 2020.

\bibitem{yang2020MPED}
Qi Yang, Siheng Chen, Yiling Xu, Sun Jun, M.~Salman Asif, and Zhan Ma.
\newblock Point cloud distortion quantification based on potential energy for
  human and machine perception.
\newblock {\em arXiv preprint arXiv:2103.02850}, 2021.

\bibitem{yang2020inferring}
Qi Yang, Zhan Ma, Yiling Xu, Zhu Li, and Jun Sun.
\newblock Inferring point cloud quality via graph similarity.
\newblock {\em IEEE Trans. Pattern Analysis and Machine Intelligence}, pages
  1--1, 2020.

\bibitem{yang2019modeling}
Qi Yang, Zhan Ma, Yiling Xu, Le Yang, Wenjun Zhang, and Jun Sun.
\newblock Modeling the screen content image quality via multiscale edge
  attention similarity.
\newblock {\em IEEE Trans. Broadcasting}, 66(2):310--321, 2019.

\bibitem{PCREG}
Haoxuan You, Yifan Feng, Ji Rongrong, and Yue Gao.
\newblock Pvnet: A joint convolutional network of point cloud and multi-view
  for 3d shape recognition.
\newblock In {\em Proceedings of the 26th ACM Int. Conf. Multimedia
  (ACMMM'18)}, pages 1310--1318, 2018.

\bibitem{SCNN}
Weixia Zhang, Kede Ma, Jia Yan, Dexiang Deng, and Zhou Wang.
\newblock Blind image quality assessment using a deep bilinear convolutional
  neural network.
\newblock {\em IEEE Trans. Circuits and Systems for Video Technology},
  30(1):36--47, 2020.

\bibitem{dnntip}
Weixia Zhang, Kede Ma, Guangtao Zhai, and Xiaokang Yang.
\newblock Uncertainty-aware blind image quality assessment in the laboratory
  and wild.
\newblock {\em IEEE Trans. Image Processing}, 30:3474--3486, 2021.

\bibitem{Zhang2021MSGRAPHSIM}
Yujie Zhang, Qi Yang, and Yiling Xu.
\newblock Ms-graphsim: Inferring point cloud quality via multiscale graph
  similarity.
\newblock In {\em Proceedings of the 29th ACM Int. Conf. Multimedia
  (ACMMM'21)}, page 1230–1238, New York, NY, USA, 2021. Association for
  Computing Machinery.

\end{thebibliography}
}

\end{document}